\documentclass[10pt]{article}

\usepackage[T1]{fontenc}
\usepackage{textcomp,fullpage}

\usepackage{helvet,courier,times,txfonts}

\usepackage{graphicx} %
\usepackage{balance}  %
\usepackage{booktabs} %
\usepackage{ccicons}  %
\usepackage{ragged2e} %

\usepackage[export]{adjustbox}

\usepackage{array}
\newcolumntype{L}[1]{>{\raggedright\let\newline\\\arraybackslash\hspace{0pt}}m{#1}}
\newcolumntype{C}[1]{>{\centering\let\newline\\\arraybackslash\hspace{0pt}}m{#1}}
\newcolumntype{R}[1]{>{\raggedleft\let\newline\\\arraybackslash\hspace{0pt}}m{#1}}

\usepackage{tikz}
\usepackage{forest}
\usetikzlibrary{positioning}
\useforestlibrary{linguistics}
\forestapplylibrarydefaults{linguistics}

\title{Principles and Practice of Explainable Machine Learning\thanks{Vaishak Belle was supported  by a Royal Society University Research Fellowship. The authors acknowledge the support received by University of Edinburgh's Bayes Centre and NatWest Group. We are especially grateful to Peter Gostev from the Data Strategy \& Innovation team as well as a  wide range of teams throughout Data \& Analytics function at NatWest Group who provided insights on industry use cases, key issues faced by financial institutions as well as on the applicability of machine learning techniques in practice.}}

\author{
\textbf{Vaishak Belle} \\
University of Edinburgh \&
Alan Turing Institute \\
{\small vaishak@ed.ac.uk} \and
\textbf{Ioannis Papantonis} \\
University of Edinburgh \\
{\small i.papantonis@sms.ed.ac.uk}
}

\begin{document}
	
\date{}

\maketitle
\begin{abstract} \it 
 Artificial intelligence (AI) provides many opportunities to improve private and public life. Discovering patterns and structures in large troves of data in an automated manner is a core component of data science, and currently drives applications in diverse areas such as computational biology, law and finance. However, such a highly positive impact is coupled with significant challenges: how do we understand the decisions suggested by these systems in order that we can trust them? In this report, we focus specifically on data-driven methods --  machine learning (ML) and pattern recognition models in particular -- so as to survey and distill the results and observations from the literature. 
 
The purpose of this report can be especially appreciated by noting that ML models are increasingly deployed in a wide range of businesses. However, with the increasing prevalence and complexity of methods, business stakeholders in the very least have a growing number of concerns about the drawbacks of models, data-specific biases, and so on. Analogously, data science practitioners are often
not aware about approaches emerging from the academic literature, or may struggle to appreciate the differences between different methods, so end up using industry standards such as SHAP. Here, we have undertaken a survey to help industry practitioners (but also data scientists more broadly)  understand the field of explainable machine learning better and apply the right tools. Our latter sections build a narrative around  a putative data scientist, and discuss how she might go about explaining her models by asking the right questions. 

From an organization viewpoint, after motivating the area broadly, we discuss the main developments, including the principles that allow us to study transparent models vs opaque models, as well as model-specific or  model-agnostic post-hoc explainability approaches.  We also briefly reflect on deep learning models, and conclude with a discussion about future research directions.

\end{abstract}

\section{Introduction}

\begin{figure}[h]
  \centering
    \includegraphics[width=.9\textwidth]{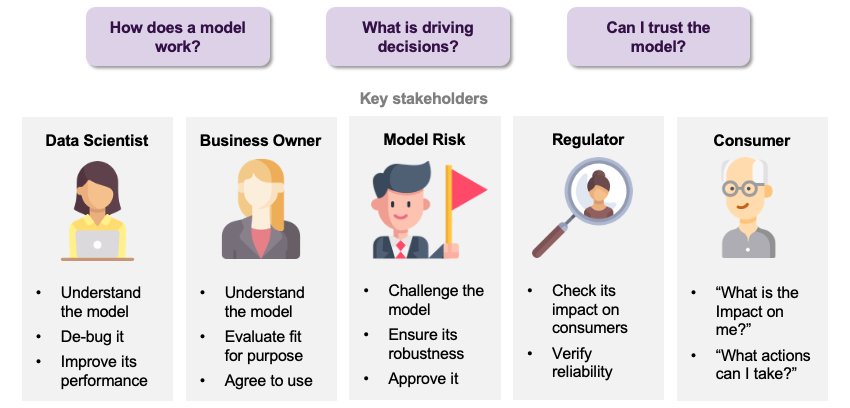}
  \caption{Concerns faced by various stakeholders}
  \label{fig:janeimages_stakeholders}
\end{figure}

Artificial intelligence (AI) provides many opportunities to improve private and public life. Discovering patterns and structures in large troves of data in an automated manner is a core component of data science, and currently drives applications in diverse areas such as computational biology, law and finance. However, such a highly positive impact is coupled with significant challenges: how do we understand the decisions suggested by these systems in order that we can trust them? Indeed, when one focuses on data-driven methods --  machine learning and pattern recognition models in particular -- the inner workings of the model can be hard to understand. In the very least, explainability can facilitate the understanding of various aspects of a model, leading to insights that can be utilized by various stakeholders, such as (cf. Figure \ref{fig:janeimages_stakeholders}): 

\begin{itemize}
    \item \textbf{Data scientists} can be benefited when debugging a model or when looking for ways to improve performance.
    \item \textbf{Business owners} caring about the fit of a model with business strategy and purpose.
    \item \textbf{Model Risk analysts} challenging the model, in order to check for robustness and approving for deployment.
    \item \textbf{Regulators} inspecting the reliability of a model, as well as the impact of its decisions on the customers.
    \item \textbf{Consumers} requiring transparency about how decisions are taken, and how they could potentially affect them.

\end{itemize}

Looking at explainability from another point of view, the developed approaches can help contribute to the following critical concerns that arise when deploying a product or taking decisions based on automated predictions: 
\begin{itemize}
\item \textbf{Correctness:} Are we confident all and only the variables of interest contributed to our decision? Are we confident spurious patterns and correlations were eliminated in our outcome? 
\item \textbf{Robustness:} Are we confident that the model is not susceptible  to minor perturbations, but if it is, is that justified for the outcome? In the presence of a missing or noisy data, are we confident the model does not misbehave? 
\item \textbf{Bias:} Are we aware of any data-specific biases that unfairly penalize groups of individuals, and if yes, can we detect and correct them? 
\item \textbf{Improvement:} In what concrete way can the prediction model be improved? What effect would additional training data or an enhanced feature space have? 
\item \textbf{Transferability:} In what concrete way can the prediction model for one application domain be applied to another application domain? What properties of the data and model would have to be adapted for this transferability? 

\item \textbf{Human comprehensibility:} Are we able to explain the model's algorithmic machinery to an expert? Perhaps even a lay person? Is that a factor for deploying the model more widely? 
\end{itemize}

The \textbf{purpose} of this report can be especially appreciated by noting that ML models are increasingly deployed in a wide range of businesses. However, with the increasing prevalence and complexity of methods, business stakeholders in the very least have a growing number of concerns about the drawbacks of models, data-specific biases, and so on. Analogously, data science practitioners are often not aware about approaches emerging from the academic literature, or may struggle to appreciate the differences between different methods, so end up using industry standards such as SHAP \cite{shap}. In this report, we have undertaken a survey to help industry practitioners (but also data scientists more broadly)  understand the field of explainable machine learning better and apply the right tools. Our latter sections particularly target how to distill and streamline questions and approaches to explainable machine learning.

\section{Development \& Contributions}

Such concerns have motivated intense activity within the community, leading to a number of involved but closely related observations. Drawing on numerous insightful surveys and perspectives (including  \cite{lipton2016mythos,arrieta2019explainable,weld2019challenge,molnar2020interpretable,doshi2017towards}) and a  large  number of available approaches, the goal of this survey is to help shed some light into the various kind of insights that can be gained, when using them. We distill concepts and strategies with the overall aim of  helping industry practitioners (but also data scientists more broadly) disentangle the different notions of explanations, as well as their intended scope of application, leading to a better understanding of the field. To this end, we first provide general perspectives on explainable machine learning that covers: notions of transparency,  criteria for evaluating explainability, as well as the type of explanations one can expect in general. We then turn to some  frameworks for summarizing the developments explainable machine learning. A taxonomic framework provides an overview of explainable ML, and the other two frameworks study certain aspects of the taxonomy. A detailed discussion on transparent vs opaque models, and model specific vs model agnostic approaches post-hoc explainability approaches follows, all of which are referred to in the taxonomic framework. Limitations and strengths of these models and approaches are discussed subsequently. 
We then turn to brief observations on explainability with respect to deep learning models. Finally, we distill these results further by building a narrative around a putative data scientist, and discuss how she might go about explaining her models. 
We conclude with some directions for future research, including the need for causality-related properties in machine learning models.

\section{Scope}

In the interest of space, we will focus on data-driven methods --  machine learning and pattern recognition models in particular -- with the primarily goal of classification or prediction by relying on statistical association. Consequently, these engender a certain class of statistical techniques for simplifying or otherwise interpreting the model at hand.

Despite this scoping, the literature is vast.\footnote{A search on Google Scholar for ``explainable machine learning" returns about one thousand results; varying search to  disjunctively include terms such as  ``interpretable'', ``artificial intelligence'', and ``explanations'', returns an even more extensive set of research papers, naturally.} Indeed,  we note that underlying concerns about human comprehensibility and generating explanations for decisions is a general  issue  in  cognitive science, social science and human psychology  \cite{miller2019explanation}. There are also various ``meta''-views on explainability, such as  maintaining an explicit model of the user \cite{chakraborti2019plan,kulkarni2019explicable}. 
Likewise, causality is expected to play a major role in explanations \cite{miller2019explanation}, but many models arising in the causality literature require careful experiment design and/or knowledge from an expert \cite{pearl2018theoretical}. They are, however, an interesting and worthwhile direction for future research, and left for concluding thoughts. Our work here primarily focuses on ``mainstream'' ML models, and the corresponding statistical explanations (however limiting they may be in a larger context) that one can extract from these models. On that note, we are not concerned with ``generating" explanations, which might involve, say, a natural language understanding component, but rather extracting an interpretation of the model's behavior and decision boundary. This undoubtedly limits the literature in terms of what we study and analyze, but it also allows us to be more comprehensive in that scope. 
For simplicity, we will nonetheless abbreviate this scoping of explainable machine learning as XAI in the report, but reiterate that the AI community takes a broader view that goes beyond (statistical) classification tasks  \cite{chakraborti2019plan,kulkarni2019explicable}. 

While we do survey and distill approaches to provide a high-level perspective, we expect the reader to have some familiarity with  classification and prediction methods. Finally, in terms of terminology, we will mostly use the term ``model" to mean the underlying machine learning technique such as random forests or logistic regression or convolutional neural networks, and use the term ``approach" and ``method''  to mean an algorithmic pipeline that is undertaken to explicitly simplify, interpret or otherwise obtain explanations from a model. If we deviate from this terminology, the context will make clear whether the entity is a machine learning or an explainability one. 

\section{Perspectives on Explainability}

Before delving into actual approaches for explainability, it is worthwhile to reflect on what are the dimensions for human comprehensibility. We will start with notions of transparency, in the sense of humans understanding the inner workings of the model. We then turn to evaluation criteria for models. We finally discuss the types of explanations that one might desire from models. It should be noted that there is considerable overlap between these notions, and in many cases, a rigorous definition or formalization is lacking and generally hard to agree on. 

\subsection{Transparency}

Transparency stands for a human-level understanding of the inner workings of the model \cite{lipton2016mythos}. We may consider three dimensions: 

\begin{itemize}
	\item \textbf{Simulatability} is the first level of transparency and it refers to a model's ability to be simulated by a human. Naturally, only models that are simple and compact fall into this category. Having said that, it is worth noting that simplicity alone is not enough, since, for example, a very large amount of simple rules would prohibit a human to calculate the model's decision simply by thought. On the other hand, simple cases of otherwise complex models, such as a neural network with no hidden layers, could potentially fall into this category.
	\item \textbf{Decomposability} is the second level of transparency and it denotes the ability to break down a model into parts (input, parameters and computations) and then explain these parts. Unfortunately, not all models satisfy this property. 
	\item \textbf{Algorithmic Transparency} is the third level and it expresses the ability to understand the procedure the model goes through in order to generate its output. For example, a model that classifies instances based on some similarity measure (such as K-nearest neighbors) satisfies this property, since the procedure is clear; find the datapoint that is the most similar to the one under consideration and assign to the former the same class as the latter. On the other hand, complex models, such as neural networks, construct an elusive loss function, while the solution to the training objective has to be approximated, too. Generally speaking, the only requirement for a model to fall into this category is for the user to be able to inspect it through a mathematical analysis. 
\end{itemize}

Broadly, of course, we may think of machine models as either being transparent or opaque/black-box, although the above  makes clear this distinction is not binary. In practice, despite the nuances, it is convention to see decision trees, linear regression, among others as simpler, transparent models, and random forests, deep learning, among others as opaque models, partly because current applications rarely use a single perceptron neural network. 
 
\subsection{Evaluation Criteria}

Although initially considered for rule extraction methods \cite{craven1999rule}, we might consider the following dimensions to evaluating models in terms of explainability: 

\begin{itemize}
	\item \textbf{Comprehensibility:} The extent to which extracted representations are humanly comprehensible, and thus touching on the dimensions of transparency considered earlier. 
	\item \textbf{Fidelity:} The extent to which extracted representations accurately capture the opaque models from which they were extracted. 
	\item \textbf{Accuracy:} The ability of extracted representations to accurately predict unseen examples.  
	\item \textbf{Scalability:} The ability of the method to scale to opaque models with large input spaces and large numbers of weighted connections. 
	\item  \textbf{Generality:} The extent to which the method requires special training regimes or restrictions on opaque models.  
\end{itemize}

We reiterate that such concepts are hard to quantify rigorously, but can nonetheless serve as guiding intuition for future developments in the area. 

\subsection{Types of explanations}

For opaque models in particular, we might consider the following types of post-hoc explanations \cite{arrieta2019explainable}: 

\begin{itemize}
	\item \textbf{Text explanations} produce explainable representations utilizing symbols, such as natural language text. Other cases include propositional symbols that explain the model's behaviour by defining abstract concepts that capture high level processes. 

\item  \textbf{Visual explanation} aim at generating visualizations that facilitate the understanding of a model. Although there are some inherit challenges (such as our inability to grasp more than three dimensions), the developed approaches can help in gaining insights about the decision boundary or the way features interact with each other. Due to this, in most cases visualizations are used as complementary techniques, especially when appealing to a non-expert audience.

\item \textbf{Local explanations} attempt to explain how a model operates in a certain area of interest. This means that the resulting explanations do not necessarily generalize to a global scale, representing the model's overall behaviour. Instead, they typically approximate the model around the instance the user wants to explain, in order to extract explanations that describe how the model operates when encountering such instances.

\item \textbf{Explanations by example} extract representative instances from the training dataset in order to demonstrate how the model operates. This is similar to how humans approach explanations in many cases, where they provide specific examples to describe a more general process. Of course, for an example to make sense, the training data has to be in a form that is comprehensible by humans, such as images, since arbitrary vectors with hundreds of variables may contain information that is difficult to uncover.

\item \textbf{Explanations by simplification} refer to the techniques that approximate an opaque model using a simpler one, which is easier to interpret. The main challenge comes from the fact that the simple model has to be flexible enough so it can approximate the complex model accurately. In most cases, this is measured by comparing the accuracy (for classification problems) of these two models.

\item \textbf{Feature relevance explanations} attempt to explain a model's decision by quantifying the influence of each input variable. This results in a ranking of importance scores, where higher scores mean that the corresponding variable was more important for the model. These scores alone may not always constitute a complete explanation, but serve as a first step in gaining some insights about the model's reasoning.  
\end{itemize}

We now turn to a distillation of the observations and techniques from the literature in the following section. We will not always be able to cover the entire gamut of dimensions considered in this section, but they do serve as a guide for the considerations to follow. 

\section{Exploring XAI}

To summarize the rapid development in explainable machine learning (XAI), we turn to five ``frameworks'' that summarize or otherwise distill the literature. 
These frameworks can be thought of as a comparative exposition and/or visualization of sorts, which help us understand: \begin{itemize}
	\item the limitations of models that may already be deployed (at least regarding explainability), 
	\item what approaches are available for explaining such models, and 
	\item what models may be considered alternatively if the application were to be redesigned with explainability in mind. 
\end{itemize}

As should be expected, there will be overlap between these frameworks.\footnote{We note that without experimental comparisons and a proper deliberation on the application domain, these frameworks purely provide an intuitive picture of model capabilities. We also note that in what follows, we make the assumption that the data is already segmented and cleaned, but it should be clear that often data pre-processing is a major step before machine learning methods can be applied. Dealing with data that has not been treated  can  affect both the applicability and the usefulness of explainability methods.}  The first two frameworks are inspired by the discussions in  \cite{arrieta2019explainable}, adapted and modified slightly for our purposes.  The third and fourth framework are based on an analysis on the current strengths and limitations of popular realizations of XAI techniques. The fifth is a ``cheat sheet'' strategy and pipeline we recommend based on  the development of numerous libraries for the analysis and interpretation of machine learning models (see, for example,  \cite{molnar2020interpretable}). 
\begin{figure}
 \includegraphics[width=.9\textwidth]{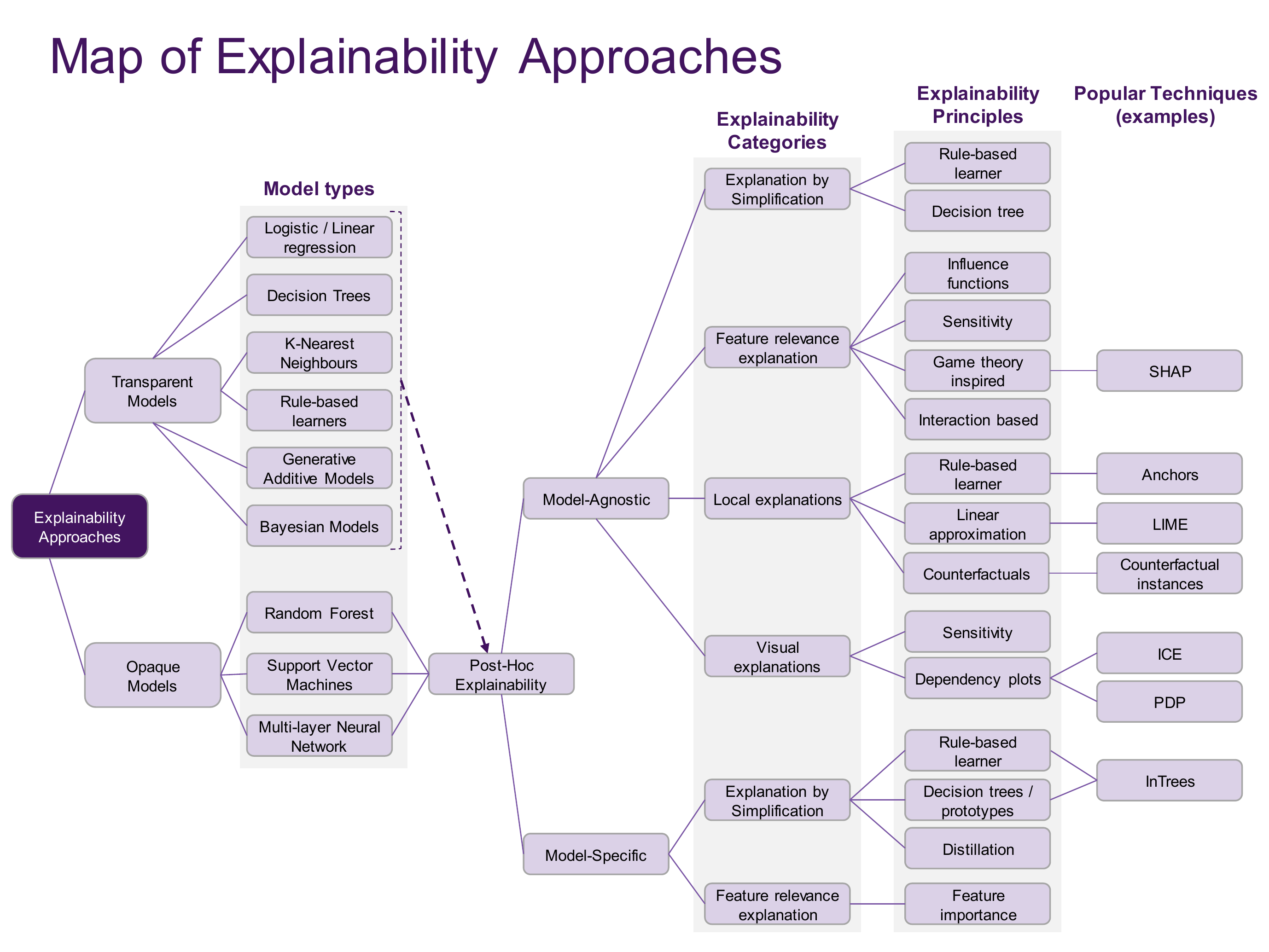}
    \caption{A taxonomic view on XAI}
    \label{fig:treeCat}
\end{figure}

\subsection{Taxonomy Framework}

In Figure \ref{fig:treeCat}, we arrange models in terms of the kinds of explainability that are enabled, to be seen as a taxonomy. The subsequent frameworks will be based on this taxonomy, and can be seen as elaborations on the distinction between transparent and opaque ML models (\textit{Transparency framework}), followed by a description of the capabilities of explainability approaches (\textit{XAI Capability framework}).

\subsection{Transparency Framework}

In Table \ref{tab:table2}, we draw a comparison between models in terms of the kinds of transparency that are enabled. This table demonstrates the correspondence between the design of various transparent ML models and the transparency dimensions they satisfy. Furthermore, it provides a summary of the most common types of explanations that are encountered when dealing with opaque models.
\begin{table*} \footnotesize 
   \centering
   \begin{tabular}{C{2cm} C{3cm} C{2cm} C{3cm} C{2cm}}
     {\small\textit{Model}} & {\small \textit{Simulatability}}
     & {\small \textit{Decomposability}}  & {\small \textit{Algorithmic Transparency}} & {\small \textit{Post-hoc}}  \\
     \midrule
     Linear/Logistic Regression & Predictors are human readable and interactions among them are kept to a minimum & Too many interactions and predictors & Variables and interactions are too complex to be analyzed without mathematical tools & Not needed \\ \vspace{10pt}

     Decision Trees & Human can understand without mathematical background &  Rules do not modify data and are understandable & Humans can understand the prediction model by traversing tree & Not needed\\ \vspace{10pt}
     
     K-Nearest Neighbors & The complexity of the model matches human naive capabilities for simulation & Too many variables, but the similarity measure and the set of variables can be analyzed & Complex similarity measure, too many variables to be analyzed without mathematical tools & Not needed\\ \vspace{10pt}
     
     Rule Based Learners & Readable variables, size of rules is manageable by a human & Size of rules is too large to be analyzed & Rules so complicated that mathematical tools are needed & Not needed\\ \vspace{10pt}
     
     General Additive Models & Variables, interactions and functions must be understandable & Interactions too complex to be simulated & Due to their complexity, variables and interactions cannot be analyzed without mathematical tools & Not needed\\ \vspace{10pt}
     
     Bayesian Models & Statistical relationships and variables should be understandable by the target audience & Relationships involve too many variables & Relationships and predictors are so complex that mathematical tools are needed & Not needed\\ \vspace{5pt}
     
     Tree Ensembles & Not applicable & Not applicable & Not applicable & Feature relevance, Model simplification\\ \vspace{5pt}
     
     Support Vector Machines & Not applicable & Not applicable & Not applicable & Feature relevance, Model simplification\\

   Multi–layer Neural Networks & Not applicable &  Not applicable &   Not applicable &  Feature relevance, Model simplification, Visualization 
   \end{tabular}
   \caption{Comparing models  on the kinds of transparency that are enabled.}~\label{tab:table2}
 \end{table*}

\subsection{XAI Capability Framework}

In Table \ref{tab:table1}, we draw a comparison between XAI approaches in terms of the type of explanations they offer, whether they are model agnostic and whether they require a transformation of the input data before the method can be applied. This summary can be utilized in order to distinguish between the capabilities of different explainability approaches, and whether the technical assumptions made for applying the approach (e.g., assumptions about independencies between variables, which is serious and limiting) is a price worth paying for the application at hand.

\renewcommand{\Huge}{}
\renewcommand{\huge}{}

\subsection{Explanation Type framework} %
\label{sub:explanation_type_framework}

In Table \ref{tab:table4}, we contrast the types of post-hoc explanations at a conceptual level: for example, what might local explanations offer in contrast to model simplification strategies?  

\subsection{Data Scientist Strategy Framework} %
\label{sub:data_scientist_strategy_framework}

In the penultimate section, we motivate a narrative for a putative data scientist, Jane, and discuss how she might go about explaining her models by asking the right questions. We recommend a simple strategy and outline sample questions that motivate certain types of explanations. \smallskip

In the following sections, we will expand on  \textit{transparent models}, followed by \textit{opaque models} and then to \textit{explainability approaches}, all of which are mentioned in the frameworks above.

\section{Transparent Models}

In this section we are going to introduce a set of models that are inherently considered to be transparent. By this, we mean that their intrinsic architecture satisfies at least one of the three transparency dimensions that we discussed in a previous section. 
\begin{itemize}
    \item \textbf{Linear\textbackslash Logistic Regression} refers to a class of models used for predicting continuous\textbackslash categorical targets, respectively, under the assumption that this target is a linear combination of the predictor variables. That specific modelling choice allows us to view the model as a transparent method. Nonetheless, a decisive factor of how a explainable a model is, has to do with the ability of the user to explain it, even when talking about inherently transparent models. In that regard, although these models satisfy the transparency criteria, they may also benefit from post-hoc explainability approaches (such as visualization), especially when non-expert audience needs to get a better understanding of the models' intrinsic reasoning.
    The model, nonetheless, has been largely applied within Social Sciences for many decades.

As a general remark, we should note that in order for the models to maintain their transparency features, their size must be limited, and the variables used must be understandable by their users.

\item \textbf{Decision Trees} form a class of models that generally fall into the transparent ML models category. They contain a set of conditional control statements, arranged in a hierarchical manner, where intermediate nodes represent decisions and leaf nodes can be either class labels (for classification problems) or continuous quantities (for regression problems). Supposing a decision tree has only a small amount of features and that its length is not prohibitively long to be memorized by a human, then it clearly falls into the class of simulatable models. In turn, if the model's length does not allow simulating it, but the features are still understandable by a human user, then the model is no longer simulatable, but it becomes decomposable. Finally, if on top of that the model also utilizes complex feature relationships, then it falls into the category of algorithmically transparent models.

Decision trees are usually utilized in cases where understandability is essential for the application at hand, so in these scenarios not overly complex trees are preferred. We should also note that apart from AI and related fields, a significant amount of decision trees' applications come from other fields, such as medicine. However, a major limitation of these models stems from their tendency to overfit the data, leading to poor generalization performance, hindering their application in cases where high predictive accuracy is desired. In such cases, ensembles of trees could offer much better generalization, but these models cannot be considered transparent anymore \footnote{Although an ensemble of a small number of decision trees could still fall under the category of transparent models, those employed in real-world applications typically consist of a large number of trees so can be seen to lose transparency properties.}.

\item \textbf{K-Nearest Neighbours (KNN)} is also a method that falls within transparent models, which deals with classification problems in a simple and straightforward way: it predicts the class of a new data point by inspecting the classes of its K nearest neighbours (where the neighbourhood relation is induced by a measure of distance between data points). The majority class is then assigned to the instance at hand. 

Under the right conditions, a KNN model is capable of satisfying any level of transparency. It should be noted, however, that this depends heavily on the distance function that is employed, as well as the model's size and the features' complexity, as in all the previous cases.

\item \textbf{Rule-based learning} is build on the intuitive basis of producing rules in order to describe how a model generates its outputs. The complexity of the resulting rules ranges from simple ``if-else'' expressions to fuzzy rules, or propositional rules encoding complex relationships between variables. As humans also utilize rules in everyday life, these systems are usually easy to understand, meaning they fall into the category of transparent models. Having said that, the exact level of transparency depends on some designing aspects, such as the the coverage (amount) and the specificity (length) of the generated rules. 

Both of these factors are at odds with the transparency of the resulting model. For example, it is reasonable to expect that a system with a very large amount of rules is infeasible to be simulated by a human. The same applies to rules containing a prohibiting number of antecedents or consequents. Including cumbersome features in the rules, on top of that, could further impede their interpretability, rendering system just algorithmically transparent.

\item \textbf{Generalized Additive Models (GAMs)} are a class of linear models where the outcome is a linear combination of some functions of the input features. The goal of these models is to infer the form of these unknown functions, which may belong to a parametric family, such as polynomials, or they could be defined non-parametrically. This allows for a large degree of flexibility, since at some applications they may take the form of a simple function, or be handcrafted to represent background knowledge, while in others they may be specified by just some properties, such as being smooth. 

These models certainly satisfy the requirements for being algorithmic transparent, at least. Furthermore, in applications where the dimensionality of the problem is small and the functions are relatively simple, they could also be considered simulatable. However, we should note that while utilizing non-parametric functional forms may enhance the models fit, it comes with a trade-off regarding its interpretability. It is also worth noting that, as with linear regression, visualization tools are often employed to communicate the results of the analysis (such as partial dependence plots \cite{doi:10.1002/sim.1501}).

\item  \textbf{Bayesian networks} refer to the designing approach where the probabilistic relationships between variables are explicitly represented using a directed graph, usually an acyclic one. Due to this clear characterization of the connection among the variables, as well as graphical criteria that examine probabilistic relationships by only inspecting the graphs topology \cite{pea}, they have been used extensively in a wide range of applications \cite{bay,ken}.

Following the above, it is clear that they fall into the class of transparent model. They can potentially fulfil the necessary prerequisites to be members of all three transparency levels, however including overly complex features or complicating graph topologies can result into them satisfying just algorithmic transparency. Research into model abstractions may be relevant to address this issue \cite{John2017-HOLPPA-12,5b737ff2f79443eb9a7905eff497b6e9}.

Owing to their probabilistic semantics, which allows conditioning and interventions, researchers have looked into ways to augment directed and undirected graphical models \cite{hmm} further  to provide explanations, although, of course, they are  already  inherently transparent in the sense described above. 
Relevant works include \cite{timmer}, where the authors propose a way to construct explanatory arguments from Bayesian models, as well as \cite{kyrimi}, where explanations are produced in order to assess the trustworthiness of a model. Furthermore, ways to draw representative examples from data have been considered, such as in \cite{kim}. 
\end{itemize}
A general remark, even when utilizing the models discussed above, is about the trade-off between complexity and transparency. Transparency, as a property, is not sufficient to guarantee that a model will be readily explainable. As we saw in the above paragraphs, as certain aspects of a model become more complex, it is not apparent how it operates internally, anymore. In these cases, XAI approaches could be used in order to explain the model's decisions, while utilizing an opaque model could also be considered.
\section{Opaque Models}

While the models we discussed  in the previous section come with appealing transparency  features, it is not always that they are among the better performing ones, at least as determined by predictive accuracy on standard (say) vision  datasets. In this section we will touch on the class of opaque models, a set of ML models which, at the expense of explainability, achieve higher accuracy utilizing complex decision boundaries.

\begin{itemize}
\item \textbf{Random Forests (RF)} were initially proposed as a way to improve the accuracy of single decision trees, which in many cases suffer from overfitting, and consequently, poor generalization. Random forests address this issue by combining multiple trees together, in an attempt to reduce the variance of the resulting model, leading to better generalization \cite{fr}. In order to achieve this, each individual tree is trained on a different part of the training dataset, capturing different characteristics of the data distribution, to obtain an aggregated prediction. This procedure results in very expressive and accurate models, but it comes at the expense of interpretability, since the whole forest is far more challenging to explain, compared to single trees, forcing the user to apply
post-hoc explainability techniques in order to gain an understanding of the decision machinery. 

\item \textbf{Support Vector Machines (SVMs)} form a class of models rooted deeply in geometrical approaches. Initially introduced for linear classification \cite{vl}, they were later extended to the non-linear case \cite{bo}, while a relaxation of the original problem \cite{co} made it suitable for real-life applications. Intuitively, in a binary classification setting, SVMs find the data separating hyperplane with the maxim margin, meaning the distance between it and the nearest data point of each class is as large as possible. Apart from classification purposes, SVMs can be applied in regression \cite{dru}, or even clustering problems \cite{ben}. While SVMs have been successfully used in a wide array of applications, their high dimensionality as well as potential data transformations and geometric motivation, make them very complex and opaque models.

\item \textbf{Multi-layer Neural Networks (NNs)} are a class of models that have been used extensively in a number of applications, ranging from bioinformatics \cite{10.1145/2649387.2649442} to recommendation systems \cite{NIPS2013_5004}, due to their state-of-the-art performance. On the other hand, their complex topology hinders their interpretability, since it is not clear how the variables interact with each other or what kind of high level features the network might has picked up. Furthermore, even the theoretical/mathematical understanding of their properties has not been sufficiently developed, rendering them virtual black-box models.

From a technical point of view, NNs are comprised of successive layers of nodes connecting the input features to the target variable. Each node in an intermediate layer collects and aggregates the outputs of the preceding layer and then produces an output on its own, by passing the aggregated value through a function (called activation function).\footnote{It is worth noting that there are a number of  options when it comes to specifying the activation function, which along with the number of the intermediate layers determine the quality of the resulting model.} In turn, these values are passed on to the next layer and this process is continued until the output layer is reached.  

An immediate observation is that as the number of layers increases, the harder it becomes to interpret the model. In contrast, an overly simple NN could even fall into the class of simulatable models. But such a simple model is of very little practical interest these days.
\end{itemize}

\section{Explainability Approaches}

In this section, we are going to review the literature and provide an overview of the various methods that have been proposed in order to produce post-hoc explanations from opaque models. The rest of the section is divided into the techniques that are especially designed for Random Forests and then we turn to  ones that are model agnostic. We focus on Random Forests owing to their popularity and to illustrate an emerging literature on model-specific explainability which often leverages technical properties of the ML model to provide a more sophisticated or otherwise customized explainability approach.

\subsection{Random Forest Explainability Approaches}

As discussed above, Random Forests are among the best performing ML algorithms, used in a wide variety of domains. However, their performance comes at the cost of explainability, so bespoke post-hoc approaches have been developed in order to facilitate the understanding of this class of models.  For tree ensembles, in general, most of the techniques found in the literature fall into either the explanation by simplification or feature relevance explanation categories. In the sequel, we will review some of the most popular approaches.

\subsubsection{Simplifying and Extracting Rules}

\begin{itemize}
    \item An attempt to simplify RFs can be found in \cite{hara}, where the authors propose a way to approximate them with a mixture of simpler models. The only requirement for the user is to specify the number of rules that the new mixture of models should contain, thereby providing a degree of freedom regarding how many rules are required to distil the model's intrinsic reasoning. Then, the resulting mixture approximates the original model utilizing only the amount of rules that the user specified. 
    
    \item Other approaches, similar in spirit, can be found in \cite{24,25}, where the objective is to approximate the RF using a single decision tree. In \cite{24}, the authors utilize a heuristic, based on information gain, in order to construct a tree that is compact enough to retain interpretability. On the other hand, the approach in \cite{25} was based on studying the asymptotic behaviour of the Gini index, in order to train a stable and accurate decision tree.
    
    \item Another approach to simplify RFs is discussed in \cite{intrees}. The main contribution of this work is proposing a methodology for extracting the more representative rules a RF has acquired. This approach remedies the fact that RFs consist of thousands of rules: by selecting only the most prominent ones, the  amount is reduced  drastically. In this case, too, the resulting rules approximate the original model, but the difference is that now rules are not learnt by a new model, but are extracted from the RF directly. Furthermore, the obtained rules can easily be combined in order to create a new rule based classifier.
    
    \item The idea above, has been explored from other perspectives as well. In \cite{10.1007}, a different method for extracting rules from a RF is proposed. In this case, a hill climbing methodology is utilized in order to construct a set of rules that approximates the original RF. This, again, leads to a significantly smaller set of rules, facilitating the model's comprehensibility.

    \item A line of research that has also been explored for producing explanations when using RFs is by providing the user with representative examples. The authors in \cite{hui} examine ways to inspect the training dataset in order to sample a number of data points that are representative members of their corresponding class. This method comes with some theoretical guarantees about the quality of the examples, while it is also adaptive, in the sense that the user specifies the number of total examples, and then the algorithm decides how many examples to sample from each class.
    
    \end{itemize}

    \subsubsection{Feature Relevance}
    \begin{itemize}
    \item Along with simplification procedures, feature relevance techniques are commonly used for  tree ensembles. One of the first approaches can be found in Breiman \cite{breiman}. His method is based on permuting the values of a feature within the dataset, and then utilizing various metrics to assess the difference in quality between the original and the newly acquired model. 
    \item The authors of \cite{Pal} develop an approach for assessing the importance of individual features, by computing how much the model's accuracy drops, after excluding a feature. Furthermore, employing this method it is possible to extract a prototypical vector of feature contributions, so we can get an idea of how important each feature is, with respect to the instances belonging in a given class. It is worth noting that extensions of this method in service  of communicating explainability  have been proposed as well, such as in \cite{floor}, where, in addition to slightly modifying the way a feature's importance is computed, graphical tools for visualizing the results are developed.
    \item A different approach on measuring a feature's importance can be found in \cite{tolomei}. The aim of this work is to examine ways to produce ``counterfactual'' data points, in the following sense: assuming a data point was classified as negative (positive), how can we generate a new data point, as similar as possible to the original one, that the model would classify as positive (negative)? The similarity metric is given by the user, so it can be application specific, incorporating expert knowledge. A by-product of this procedure is that by examining the extent to which a feature was modified, we get an estimate of its importance, as well as the new counterfactual data point.
    
    \item In a somewhat different, yet relevant, approach the authors in \cite{imp} develop a series of metrics assessing the importance of the model's features. Apart from standard importance scores, they also discuss how to answer more complex questions, such as what is the effect on the model's accuracy, when using only a subset of the original features, or which subsets of features interact together.
    
    \item Other ways to identify a set of important features can be found in the literature, as well. The authors in \cite{auret} propose a way to determine a threshold for identifying important features. All features exceeding this threshold are deemed important, while those that do not are discarded as unnecessary. Following this approach, apart from having a vector with each feature's importance, a way to identify the irrelevant ones is also provided. In addition, graphical tools to communicate the results to a non-expert audience are discussed. 
\end{itemize}

\begin{table*}[t] \footnotesize 
   \centering %
   \begin{tabular}{C{2cm} C{1cm} C{2cm} C{1cm} C{1.5cm} C{1cm} C{1.5cm} C{1.5cm} C{1cm} C{1cm}}
     {\small\textit{XAI method}} & {\small \textit{Swapping}}
     & {\small \textit{Explanation}}  & {\small \textit{Model agnostic}}  & {\small \textit{Categorical/Continuous features}} & {\small \textit{Intermediate transformation}} & {\small \textit{Independent features}} & {\small \textit{Shapley values}} & {\small \textit{Examples}} \\
     \midrule
     KernelSHAP \cite{shap} & No &  Feature relevance & Yes & Both & Yes & Yes & Yes & No \\
     TreeSHAP \cite{shap} & Yes & Feature relevance & No & Both & Yes & No & Yes & No\\
     LIME \cite{lime} & Yes &  Simplification & Yes & Both & Yes & No & Not necessarily & No\\ 
     Anchors \cite{AAAI1816982} & Yes & Simplification & Yes & Both & No & No & No & No\\
     QII \cite{datta} & Yes &  Feature relevance & Yes & Both & Yes & No & Not necessarily & No\\
     CNF rules \cite{boolean} & Yes &  Simplification & Yes & Categorical & No & No & No & No\\
     Influence function \cite{influence} & Yes &  Feature relevance & Yes & Both & No & No & No & Yes\\ 
    ASTRID \cite{interpreting} & Yes &  Feature relevance & Yes & Both & No & No & No & No\\ 
    Distilation \cite{distil} & Yes &  Simplification & Yes & Both & No & No & No & No\\ 
    Counterfactual \cite{counterfactual} & Yes &  Local & Yes & Both & No & No & No & Yes\\
   InTrees \cite{intrees} & Yes &  Simplification & No & Both & No & No & No & No\\ 
    
    Prototypes \cite{hui} & Yes &  Simplification & No & Both & No & No & No & Yes\\
    Feature tweaking \cite{tolomei} & Yes &  Feature relevance & No & Both & No & No & No & Yes\\
   \end{tabular}
   \caption{Comparing XAI methods}~\label{tab:table1}
 \end{table*}

\subsection{Model-agnostic Explainability Approaches}

Model-agnostic techniques are designed having the purpose of being generally  applicable, in mind. They have to be flexible enough, so they do not depend on the intrinsic architecture of a model, thus operating solely on the basis of relating the input of a model to its outputs. Arguably, the most prominent explanation types in this class are model simplification, feature relevance, as well as visualizations.

% \begin{itemize}
% \item \textit

\subsubsection{Explanation by simplification} % (fold)
\label{ssub:subsubsection_name}

% subsubsection subsubsection_name (end){} 

%

Arguably the most popular is the technique of Local Interpretable Model-Agnostic Explanations (LIME) \cite{lime}. LIME approximates an opaque model locally, in the surrounding area of the prediction we are interested in explaining, building either a linear model or a decision tree around the predictions of an opaque model, using the resulting model as a surrogate in order to explain the more complex one. Furthermore, this approach requires a transformation of the input data to an ``interpretable representation'', so the resulting features are understandable to humans, regardless of the actual features used by the model (this is termed as ``intermediate transformation'', in Table \ref{tab:table1}). 

A similar technique, called anchors, can be found in \cite{AAAI1816982}. Here the objective is again to approximate a model locally, but this time not by using a linear model. Instead, easy to understand ``if-then'' rules that anchor the model's decision are employed. The rules aim at capturing the essential features, omitting the rest, so it results in more sparse explanations.

G-REX \cite{grex} is an approach first introduced in genetic programming, in order to extract rules from data, but later works have expanding its score, rendering capable of addressing explainability \cite{Johansson,truth}.

Another approach is introduced in \cite{boolean}, where the authors explore a way to learn rules in either Conjunctive Normal Form (CNF) or Disjunctive Normal Form (DNF). Supposing that all variables are binary, then the algorithm builds a classification model that attempts to explain the complex model's decisions utilizing only such propositional rules. Such approaches have the extra benefit of resulting in a set of symbolic rules that are explainable by default, as well as can be utilized as a predictive model, themselves.

Another perspective in simplification is introduced in \cite{10.1145/3077257.3077271}. In this work, the objective is to approximate an opaque model using a decision tree, but the novelty of the approach lies on partitioning the training dataset in similar instances, first. Following this procedure, each time a new data point is inspected, the tree responsible for explaining similar instances will be utilized, resulting in better local performance. Additional techniques to construct rules explaining a model's decisions can be found in \cite{7738872,turner2016model}.

In similar spirit, the authors of \cite{extraction}  formulate model simplification as a model extraction process by approximating a complex model using a transparent one. The proposed approach utilizes the predictions of a black-box model to build a (greedy) decision tree, in order to inspect this surrogate model to gain some insights about the original one.  Simplification is approached from a different perspective in \cite{distil}, where an approach to distill and audit black box models is presented. This is a two-part process, comprising of a distillation approach, as well as a statistical test. So, overall, the approach provides a way to inspect whether a set of variables is enough to recreate the original model, or if extra information is required in order to achieve the same accuracy. 

There has been considerable recent development in the so-called counterfactual explanations  \cite{counterfactual}. Here, the objective is to create instances as close as possible to the instance we wish to explain, but such that the model classifies the new instance in a different category. 
By inspecting this new data point and comparing it to the original one we can gain insights on what the model considers as minimal changes to the original data point, so as to change its decision. 
A simple example is the case of an applicant who was denied his loan application, and the explanation might say that had he had a permanent contract with his current employer, the loan would be approved.

\subsubsection{Feature relevance}

One of the most popular contributions here, and in XAI in general, is that of SHAP (SHapley Additive exPlanations)  \cite{shap}. The objective in this case is to build a linear model around the instance to be explained, and then interpret the coefficients as the feature's importance. This idea is similar to LIME, in fact LIME and SHAP are closely related, but SHAP comes with a set of nice theoretical properties. Its mathematical basis  is rooted in coalitional Game Theory, specifically on Shapley values \cite{shapley}. Roughly, the Shapley value of a feature is its average expected marginal contribution to the model's decision, after all possible combinations have been considered. However, the dimensionality of many complex real-world  applications renders the calculation of these values infeasible, so the authors in  \cite{shap}  simplify the problem by making various assumptions, such as independency among the variables. Arguably, this is a strong assumption that can affect the quality of the resulting Shapley values. Other issues exist as well, for example while in its formulation all  possible orderings of the variables are considered, in practice this is infeasible, so the ordering of the variable affects the computation of the Shapley values. (In Table \ref{tab:table1}, for example, we use the term ``swapping'' to refer to whether a method is influenced by the features' ordering.) 

Similar in spirit,  in \cite{game},  the authors propose to measure a feature's importance using its Shapley value, but the objective function, as well as the optimization approach, is not the same as in SHAP. A different strategy is considered in \cite{datta}, where a broad variety of measures are presented to tackle the quantification of the degree of influence of inputs on the outputs. The proposed QII (Quantitative Input Influence) measures account for correlated inputs, which  quantifies the influence by estimating the change in performance when using the original data set versus when using one where the feature of interest is replaced by a random quantity.  

Another approach that is based on random feature permutations can be found in \cite{peek}. In this work, a methodology for randomizing the values of a feature, or a group of features, is introduced, based on the difference between the model's behaviour when making predictions for the original dataset and when it does the same for the randomized version. This process facilitates the identification of important variables or variable interactions the model has picked up.

Additional ways to assess the importance of a feature can also be found, such as the one in \cite{adebayo2016iterative}. The authors introduce a methodology for computing feature importance, by transforming each feature in a dataset, so the result is a new dataset where the influence of a certain feature has been removed, meaning that the rest of the attributes are orthogonal to it. By using several modified datasets, the authors develop a measure for calculating a score, based on the difference in the model's performance across the various datasets.

Different from the above threads, in \cite{cortez}, the authors extend existing SA (Sensitivity Analysis) approaches in order to design a Global SA  method. The proposed methodology is also paired with  visualization tools to facilitate communicating the results. Likewise, the work in \cite{interpreting} presents a method (ASTRID) that aims at identifying which attributes are utilized by a classifier in prediction time. They approach this problem by looking for the largest subset of the original features so that if the model is trained on this subset, omitting the rest of the features, the resulting model would perform as well as the original one. In \cite{influence}, the authors use influence functions to trace a model’s prediction back to the training data, by only requiring an oracle version of the model with access to gradients and Hessian-vector products. Finally, another way to measure a data point's influence on the model's decision comes from deletion diagnostics \cite{doi:10.1080/00401706.1977.10489493}. The difference this time is that this approach is concerned with measuring how omitting a data point from the training dataset influences the quality of the resulting model, making it useful for various tasks, such as model debugging.

\subsubsection{Visual explanations}

Some popular approaches to visualizations can be found in \cite{cortez}, where an array of various plots are presented. Additional techniques are discussed in \cite{CORTEZ20131}, where some new SA approaches are introduced. Finally, \cite{goldstein,friedman2001} presents ICE (Individual Conditional Expectation) and PD (Partial Dependence) plots, respectively. The former, operates on instance level, depicting the model's decision boundary as a function of a single feature, with the rest of them staying fixed. The latter, again plots the model's decision boundary as a function of a single feature, but this time the remaining features are averaged out, so this shows the average effect. There is an interesting relationship between these two plots, as averaging the ICE plots of each instance of a dataset, yields the corresponding PD plot.

Along with the three frameworks, the above exposition covers the main observations and properties of XAI trends. 

\section{Brief Overview of  Deep Learning Models}

In this section we provide a brief summary of XAI approaches that have been developed for deep learning (DL) models, specifically multi-layer neural networks (NNs). NNs are highly expressive computational models, achieving state-of-the-art performance in a wide range of applications. Unfortunately, their architecture and learning regime corresponds to a complex computational pipeline, so they do not satisfy any level of transparency, at least when we go beyond simple models, such as single layer perceptron as mentioned previously, although, of course such models do not fall within ``deep'' learning. This has led to the development of NN-specific XAI methods, utilizing their specific topology. The majority of these methods fall into the category of either \textit{model simplification} or \textit{feature relevance}. 

In \textit{model simplification}, rule extraction is one of the most prominent approaches. Rule extraction techniques that operate on a neuron-level rather than the whole model are called decompositional. \cite{kulluk} propose a method for producing if-else rules from NNs, where model training and rule generation happen at the same time. CRED \cite{cred} is a different approach that utilizes decision trees in order to represent the extracted rules. KT \cite{kt} is a related algorithm producing if-else rules, in a layer by layer manner. DeepRED \cite{deepred} is one of the most popular such techniques, extending CRED. The proposed algorithm has additional decision trees as well as intermediate rules for every hidden layer. It can be seen as a divide and conquer method aiming at describing each layer by the previous one, aggregating all the results in order to explain the whole network.

On the other hand, when the internal structure of a NN is not taken into account, the corresponding methods are called pedagogical. That is, approaches that treat the whole network as a black-box function and do not inspect it at a neuron-level in order to explain it. TREPAN \cite{trepan} is such an approach, utilizing decision trees as well as a query and sample approach. Saad and Wunsch \cite{SAAD} have proposed an algorithm called HYPINV, based on a network inversion technique. This algorithm is capable of producing rules having the form of the conjunction and disjunction of hyperplanes. Augusta and Kathirvalavakumar \cite{auga} have introduced the RxREN algorithm, employing reverse engineering techniques in order to analyse the output and trace back the components that cause the final result. 

Combining the above approaches leads to eclectic rule extraction techniques. RX \cite{HRUSCHKA2006384} is such a method, based on clustering the hidden units of a NN and extracting logical rules connecting the input to the resulting clusters. An analogous eclectic approach can be found in \cite{KAHRAMANLI20091513}, where the goal is to generate rules from a NN, using so-called  artificial immune system (AIS) \cite{dip} algorithms.

Apart from rule extraction techniques, other approaches have been proposed in order to interpret the decisions of NNs. In \cite{mimic}, the authors introduce \textit{Interpretable Mimic Learning}, which builds on model distillation ideas, in order to approximate the original NN with a simpler, interpretable model. The idea of transferring knowledge from a complex model (the \textit{teacher}) to a simpler one (the \textit{student}) been explored in other works, for example \cite{44873,Bucila2006ModelC,NIPS2019_9151}.

An intuitive observation about NNs is that as the number of layers grows larger, developing model simplification algorithms gets progressively more difficult. Due to this, \textit{feature relevance} techniques have gained popularity in recent years. In \cite{Kindermans}, the authors propose ways to estimate neuron-wise signals in NNs. Utilizing these estimators they present an approach to superposition neuron-wise explanations in order to produce more comprehensive explanations.

In \cite{MONTAVON} a way to decompose the prediction of a NN is presented. To this end, a neuron's activation is decomposed and then its score is backpropagated to the input layer, resulting in a vector containing each feature's importance. 

DeepLIFT \cite{pmlr-v70} is another way to assign importance scores when using NNs. The idea behind this method is to compare a neuron's activation to a reference one and then use their difference to compute the importance of a feature.

Another popular approach can be found in \cite{10.5555}, where the authors present Integrated Gradients. In this work, the main idea is to examine the model's behaviour when moving along a line connecting the instance to be explained with a baseline instance (serving the purpose of a ``neutral'' instance). Furthermore, this method comes with some nice theoretical properties, such as \textit{completeness} and \textit{symmetry preservation}, that provide assurances about the generated explanations.

\section{Views and Suggestions}

XAI is a broad and relatively new branch of ML, which, in turn, means that there is still some ambiguity regarding the goals of the resulting approaches. The approaches presented in this survey are indicative of the range of the various explainability angles that are considered within the field. For example, feature relevance approaches provide insights by measuring and quantitatively ranking the importance of a feature, model simplification approaches construct relatively simple models as proxies for the opaque ones, while visual explanations inspect a model's inner understanding of a problem through graphical tools. At this point we should note that choosing the right technique for the application at hand depends exactly at the kind of insights the user would like to gain, or perhaps the kind of explanations he/she is more comfortable interpreting. 

In applications were explainability is of utmost importance, it is worth considering using a transparent model. The downside of this, is that these models often compromise performance for the sake of explainability, so it is possible that the resulting accuracy hinders their employment in crucial real-world applications. 

In cases where maintaining high accuracy is a driving factor, too, employing an opaque model and pairing it with some XAI techniques, instead of using a transparent one, is probably the most reasonable choice. Subsequently, identifying the right technique for explaining the resulting model is the next step in the quest to understand its internal mechanisms. Each of them comes with its own strong points, as well as limitations. More specifically:

\begin{itemize}
    \item Local explanations approximate the model in a narrow area, around a specific instance of interest. They offer information about how the model operates when encountering inputs that are similar to the one why are interested in explaining. This information can attain various forms, such as importance scores or rules. Of course, this means that the resulting explanations do not necessarily reflect the model's mechanism on a global scale. Other limitations arise when considering the inherent difficulty to define what a local area means in a high dimensional space. This could also lead to cases where slightly perturbing a feature's value results in significantly different explanations.
    
    \item Representative examples allows the user to inspect how the model perceives the elements belonging in a certain category. In a sense, they serve as prototype data points. In other related approaches, it is possible to trace the model's decision back to the training dataset and uncover the instance that influenced the model's decision the most. Deletion diagnostics also fall into this category, quantifying how the decision boundary changes when some training datapoints are left out. The downside of utilizing examples is that they require human inspection in order to identify the parts of the example that distinguish it from the other categories. 
    
    \item Feature relevance explanations aim at computing the influence of a feature in the model's outcome. This could be seen as an indirect way to produce explanations, since they only indicate a feature's individual contribution, without providing information about feature interactions. Naturally, in cases where there are strong correlations among features, it is possible that the resulting scores are counterintuitive. On the other hand, some of these approaches, such as SHAP, come with some nice theoretical properties (although in practice they might be violated \cite{kumar2020problems,merrick2019explanation}). 
    
    \item Model simplification comes with the immediate advantage and flexibility of allowing to approximate an opaque model using a simpler one. This offers a wide range of representations that can be utilized, from simple ``if-then'' rules to fitting surrogate models. This way explanations can be adjusted to best fit a particular audience. Of course, there are limitations as well, with perhaps the most notable one being the quality of the approximation. Furthermore, usually, it is not possible to quantitatively assess it, so empirical demonstrations are needed to demonstrate the goodness of the approximation.
    
    \item Visualizations provide for a way to utilize graphical tools in order to inspect some aspects of a model, such as its decision boundary. In most cases they are relatively easy to understand for both technical and non technical audiences. However, when resorting to visualizations, many of the proposed approaches make assumptions about the data (such as independence) that might not hold for the particular application, perhaps distorting the results.
    
\end{itemize}

Overall, we summarize some of the salient properties to consider in Table \ref{tab:table4}.

Taking a close look at the various kinds of explanations discussed above, makes clear that each of them addresses a different aspect of explainability. This means that there is no approach suitable for each and every scenario. This is in tune with how humans perceive explainability as well, since we know that there is not a single question whose answer would be able to communicate all the information needed to explain any situation. Most of the times, one would have to ask multiple questions, each one dealing with a different aspect of the situation in order to obtain a satisfactory explanation.

The same approach should be utilized when inspecting the reasoning of ML models. Relying on only one technique will only give us a partial picture of the whole story, possibly missing out important information. Hence, combining multiple approaches together provides for a more cautious way to explain a model. 

At this point we would like to note that there is no established way of combining techniques (in a pipeline fashion), so there is room for experimenting and adjusting them, according to the application at hand. Having said that, we think that a reasonable base case could look like this:
\begin{itemize}
    \item If explainability is essential for the application, first try transparent models. 
    \item If it doesn't perform well, and particularly if the complexity of the model is escalating and you lose the explainability benefit, use an opaque one.
    \item  Employ an feature relevance method to get the an instance-specific estimate of each feature's influence.
    \item  A model simplification approach could be used to inspect whether the important features, will turn out to be important on a global scale, too.
    \item A local explanation approach could shed light into how small perturbations affect the model's outcome, so pairing that with the importance scores could facilitate the understanding of a feature's significance. 
    \item A visualization technique to plot the decision boundary as a function of a subset of the the important features, so we can get a sense of how the model's predictions change.

\end{itemize}

\begin{table*} \footnotesize 
   \centering
   \begin{tabular}{C{2cm} C{6cm} C{6cm} }
     {\small\textit{Explanation}} & {\small \textit{Advantages}}
     & {\small \textit{Disadvantages}}   \\
     \midrule
     
     Local explanations & Explains the model's behaviour in a local area of interest.  Operates on instance-level explanations. & Explanations do not generalize on a global scale. Small perturbations might result in very different explanations. Not easy to define locality. Some approaches face stability issues.  \\  \hline

Examples  & Representative examples provide insights about the model's internal reasoning. Some of the algorithms uncover the most influential training data points that led the model to its predictions.&  Examples require human inspection. They do not explicitly state what parts of the example influence the model. \\  \hline

Feature relevance & They operate on an instance level, calculating the importance of each feature in the model's decision. A number of the proposed approaches come with appealing theoretical guarantees. & They are sensitive in cases where the features are highly correlated. In many cases the exact solutions are approximated, leading to undesirable side effects, such as the ordering affecting the outcome. \\ \hline
Simplification & Simple surrogate models explain the opaque ones. Resulting explanations, such as rules, are easy to understand. & Surrogate models may not approximate the original models well. Surrogate models come with their own limitations. \\ \hline
Visualizations & Easier to communicate to non technical audience. Most of the approaches are intuitive and not hard to implement. & There is an upper bound on how many features we can consider at once. Humans need to inspect the resulting plots in order to produce explanations.  \\ \vspace{10pt}

   \end{tabular}
   \caption{Advantages and disadvantages of the various kinds of explanations.}~\label{tab:table4}
 \end{table*}

{}

\section{Jane, the Data Scientist}

In this section we will discuss a concrete example of how a data scientist could apply the insights gained so far, in a real-life scenario. To this end, we would like to introduce Jane, a data scientist whose work is on building ML models for loan approvals. As a result, she wold like to consider things like the likelihood of default given some parameters in a credit decision model. 

\begin{figure}[h]
  \centering
    \includegraphics[width=.9\textwidth]{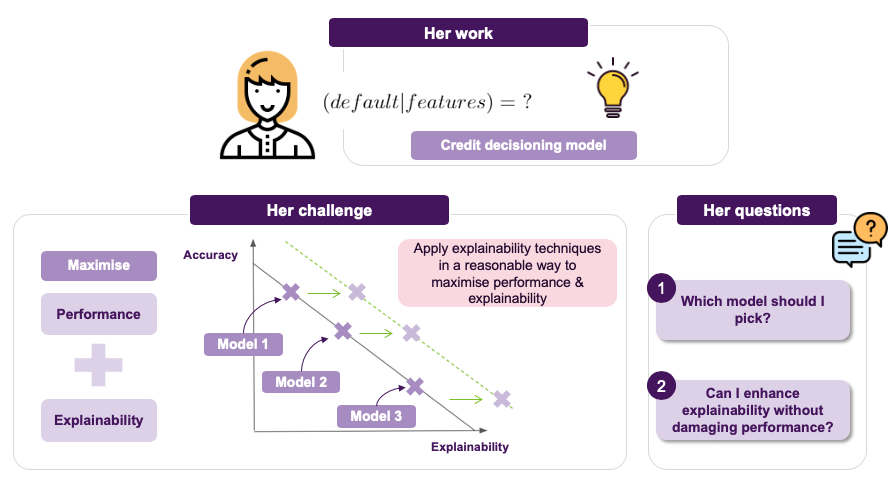}
  \caption{Jane's agenda and challenge: which model offers the best trade-off in terms of accuracy vs explainability?}
  \label{fig:janeimages_janequestions}
\end{figure}

\begin{figure}[h]
  \centering
    \includegraphics[width=.9\textwidth]{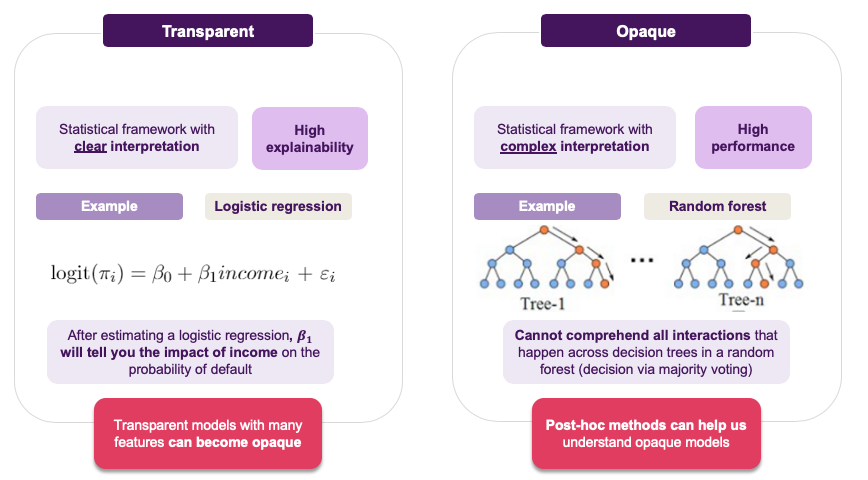}
  \caption{Jane's choices: should she go for a transparent model or an opaque one?}
  \label{fig:janeimages_choices}
\end{figure}

\begin{figure}[h]
  \centering
    \includegraphics[width=.9\textwidth]{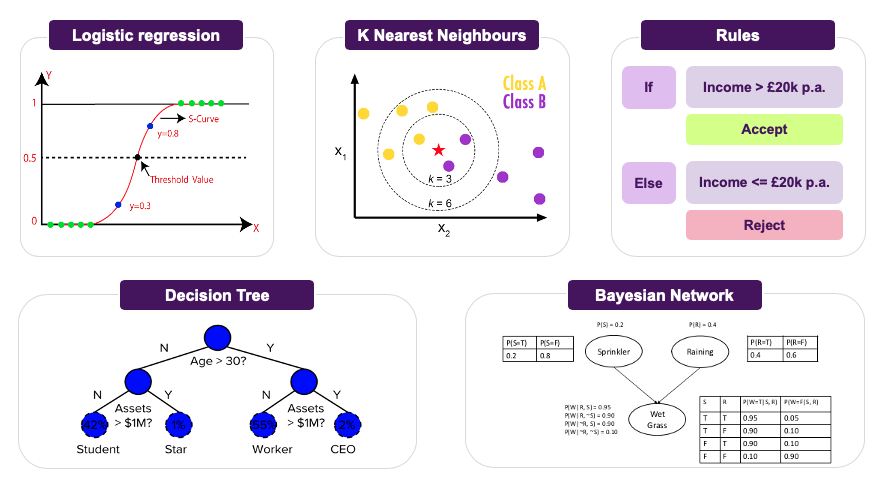}
  \caption{Some popular transparent models.}
  \label{fig:janeimages_transparent}
\end{figure}

\begin{figure}[h]
  \centering
    \includegraphics[width=.9\textwidth]{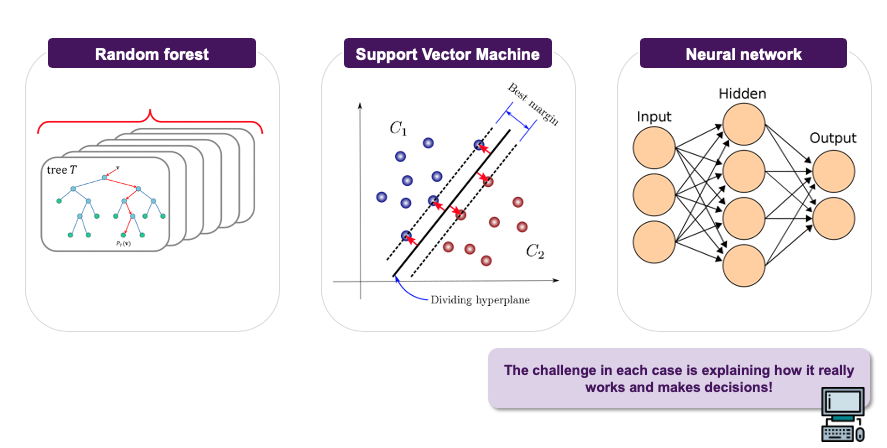}
  \caption{Some popular opaque models.}
  \label{fig:janeimages_opaque}
\end{figure}

Jane's current project is to employ a model to assess whether a loan should be approved, that maximizes performance while also maintaining explainability (cf. Figure \ref{fig:janeimages_janequestions}).\footnote{Note that this informal view encourages a notional plot of explainability versus accuracy, as is common in informal discussions on the challenge of XAI \cite{gunning2017explainable,weld2019challenge}. However, this informal view has been criticized \cite{rudin2019stop} as being misleading. Since we are concerned primarily with mainstream ML models and the interpretability that emerges when applying statistical analysis to such models, we will continue using this notional idea for the sake of simplicity.} This leads to the challenge of achieving an optimal trade-off  between these two things. Broadly, we can think of two possible choices for Jane (cf. Figure \ref{fig:janeimages_choices}):

\begin{itemize}
    \item She can go for transparent models (cf. Figure \ref{fig:janeimages_transparent} for popular choices), resulting in a clear interpretation of the decision boundary, allowing for immediately interpreting how a decision is made. For example, if using logistic regression, the notion of defaulting can seen as a weighted sum of features, so a feature's coefficient will tell you this feature's impact on defaulting. 
    \item Otherwise, she can go for an opaque model (cf. Figure \ref{fig:janeimages_opaque} for popular choices), which usually achieve better performance and generalizability than their transparent counterparts. Of course, the downside is that in this case is it will not be easy to interpret the model's decisions.
\end{itemize}

\begin{figure}[h]
  \centering
    \includegraphics[width=.9\textwidth]{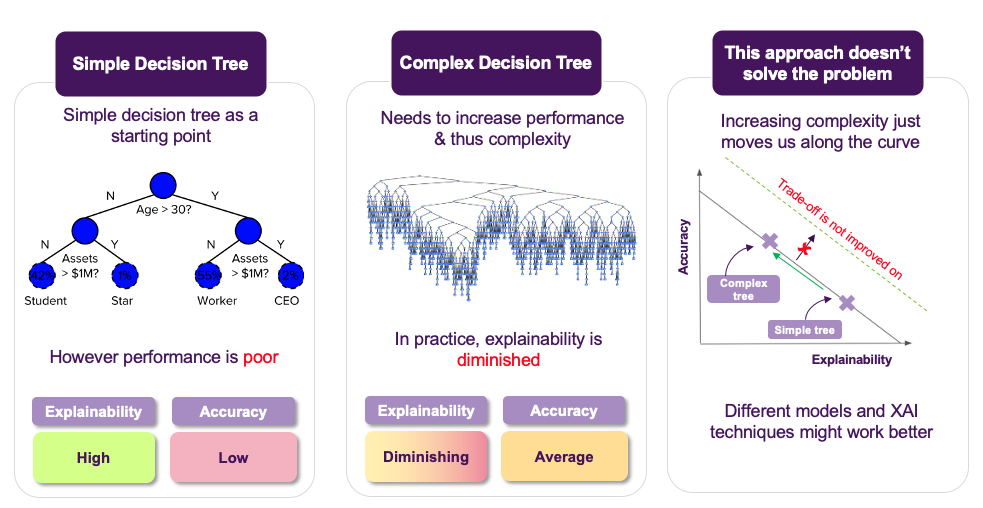}
  \caption{As transparent models become increasingly complex they may lose their explainability features. The primary goal is to maintain a balance between explainability and accuracy. In cases where this is not possible, opaque models paired with post hoc XAI approaches provide an alternative solution.}
  \label{fig:janeimages_switch}
\end{figure}

\begin{figure}[h]
  \centering
    \includegraphics[width=.9\textwidth]{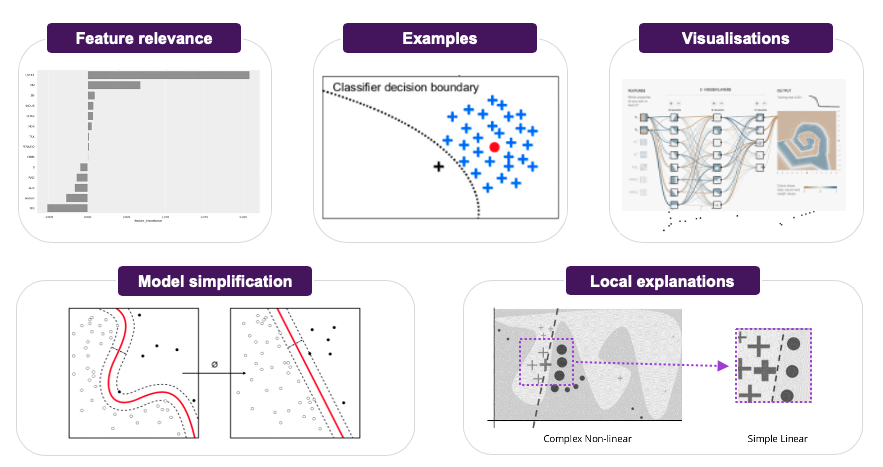}
  \caption{Popular choices for post-hoc explanations.}
  \label{fig:janeimages_posthocoptions}
\end{figure}

Jane decides to give various transparent models a try, but the resulting accuracy was not satisfactory, so she resorts to opaque models (cf. Figure \ref{fig:janeimages_switch}). She again tries various candidates and she finds out that Random Forests achieve the best performance among them, so this is what she will use. After training the model, the next step is to come up with ways that could help her explain how the model operates to the stakeholders (cf. Figure \ref{fig:janeimages_posthocoptions} for popular choices).

\begin{figure}[h]
  \centering
    \includegraphics[width=.9\textwidth]{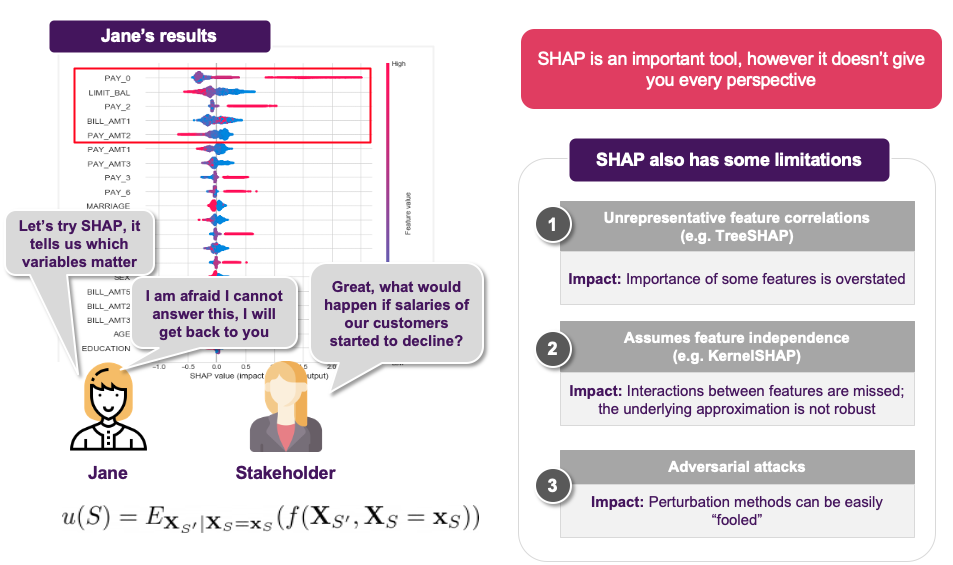}
  \caption{Jane decides to use SHAP, but cannot resolve all of the stakeholder's questions. Its also worth noting that although SHAP is an important method for explaining opaque models,  users should be aware of its limitations, often arising from either the optimization objective or the underlying approximation. 
  }
  \label{fig:janeimages_shap}
\end{figure}

The first thing that came to Jane's mind was to utilize one of the most popular XAI techniques, SHAP. She goes on applying it in order to explain a specific decision made by the model. She computes the importance of each feature and shares it with the stakeholders to help them understand how the model operates. However, as the discussion progresses, a reasonable question comes up (Figure \ref{fig:janeimages_shap}): could it be that the model relies heavily on an applicant's salary, for example, missing other important factors? How would the model perform on instances where applicants have a relatively low salary? For example, assuming that everything else in the current application was held intact, what is the salary's threshold that differentiates an approved from a rejected application?

\begin{figure}[h]
  \centering
    \includegraphics[width=.9\textwidth]{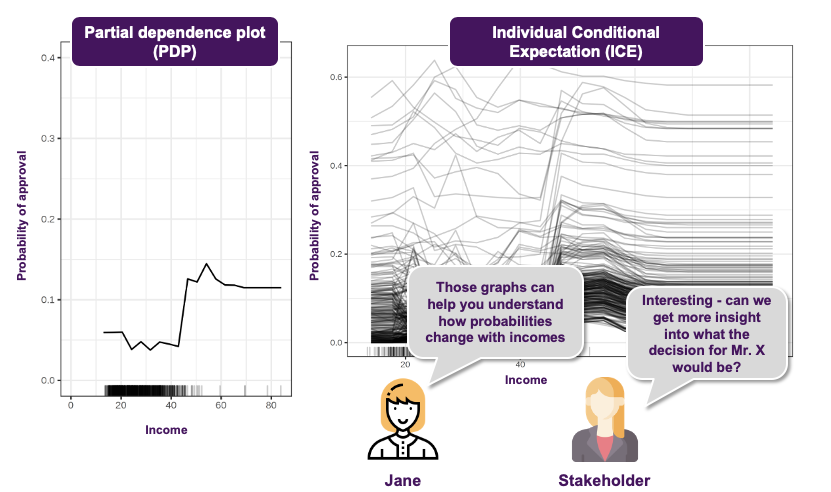}
  \caption{Visualizations can facilitate understanding the model's reasoning, both on an instance and a global level. Most of these approaches make a set of assumptions, so choosing the  appropriate one depends on the application.}
  \label{fig:janeimages_pdp}
\end{figure}

These questions cannot been addressed using SHAP, since they refer to how the model's predictive behaviour would change, where SHAP can only explain the instance at hand, so Jane realises that she will have to use additional techniques to answer these questions. To this end, she decides to employ Individual Conditional Expectation (ICE) plots, to inspect the model's behaviour for a specific instance, where everything except salary is held constant, fixed to their observed values, while salary is free to attain different values. She could also compliment this technique using Partial Dependence Plots (PDPs) to plot the model's decision boundary as a function of the salary, when the rest of the features are averaged out. This plot allows her to gain some insights about the model's average behavior, as the salary changes (Figure \ref{fig:janeimages_pdp}). 

\begin{figure}[h]
  \centering
    \includegraphics[width=.9\textwidth]{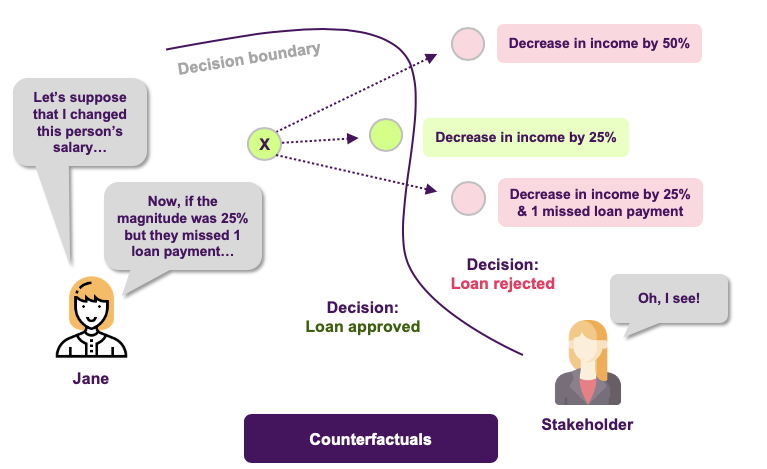}
  \caption{Counterfactuals produce a hypothetical instance, representing a minimal set of changes of the original one, so the model classifies it in a different category.}
  \label{fig:janeimages_counterfactuals}
\end{figure}

Jane discusses her new results with the stakeholders, explaining how these plots provide answers to the questions that were raised, but this time there is a new issue to address. In the test set there is an application that the model rejects, which comes contrary to what various experts in the bank think should have happened. This leaves the stakeholders in question of why the model decides like that and whether a slightly different application would have been approved by the model. Jane decides to tackle this using counterfactuals, which inherently convey a notion of ``closeness'' to the actual world. She applies this approach and she finds out that it was the fact that the applicant had missed one payment that led to this outcome, and that had he/she missed none the application would had been accepted (Figure \ref{fig:janeimages_counterfactuals}).

\begin{figure}[h]
  \centering
    \includegraphics[width=.9\textwidth]{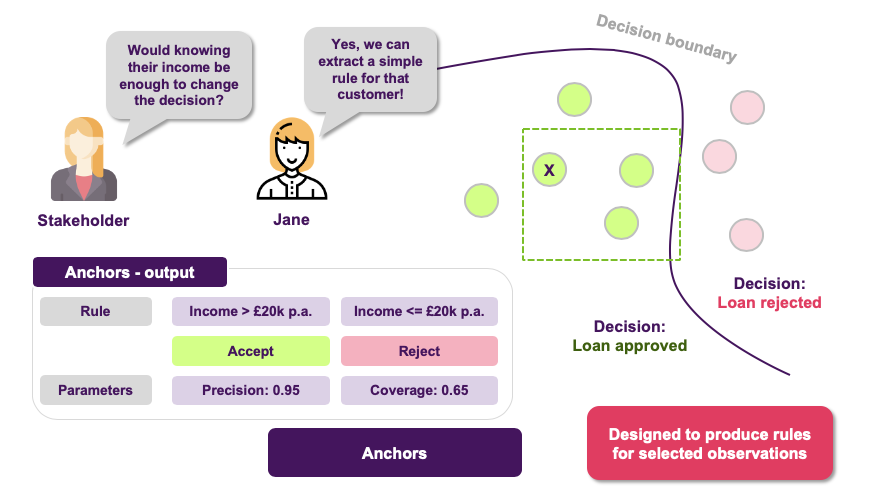}
  \caption{Local explanations as rules. High precision means that the rule is robust and that similar instances will get the same outcome.
High coverage means that large number  of the points satisfy the rule's premises, so the rule ``generalizes"  better.
}
  \label{fig:janeimages_anchors}
\end{figure}

\begin{figure}[h]
  \centering
    \includegraphics[width=.9\textwidth]{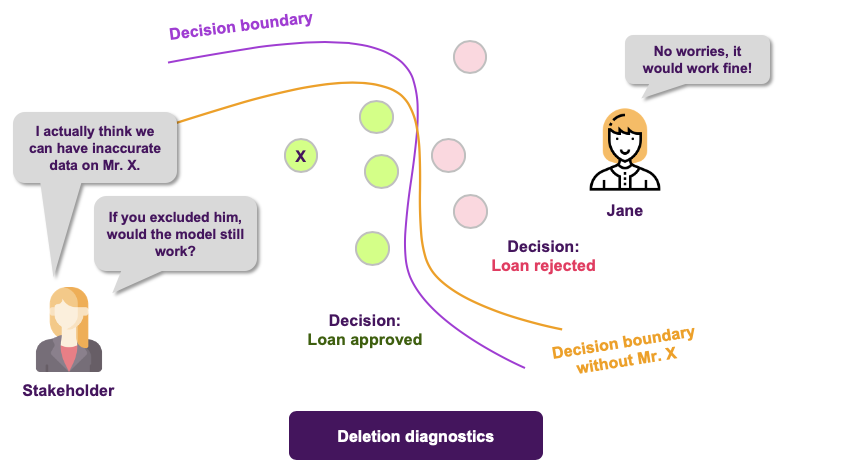}
  \caption{The quality of a ML model is vastly affected by the quality of the data it is trained on. Finding influential points that can, for example, alter the decision boundary or encourage the model to take a certain decision, contributes in having a more complete picture of the model's reasoning.}
  \label{fig:janeimages_deletion}
\end{figure}

The stakeholders think this is a reasonable answer, but now that they saw how influential the number of missed payments was, they feel that it would be nice to be able to extract some kind of information explaining how the model operates for instances that are similar to the one under consideration, for future reference. 

\begin{figure}[h]
  \centering
    \includegraphics[width=.9\textwidth]{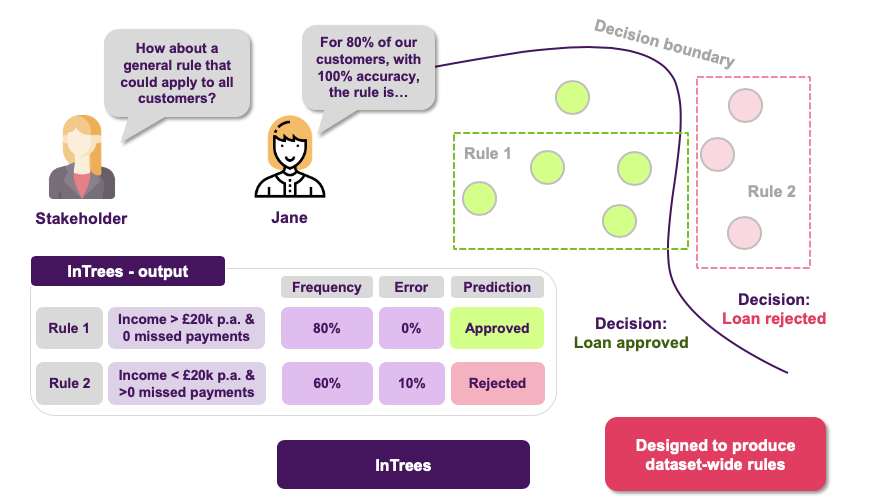}
  \caption{Extracting rules from a random forest. Frequency of a rule is defined as the proportion of data instances satisfying the rule condition. The frequency measures the popularity of the rule.
Error of a rule is defined as the number of incorrectly classified instances determined by the rule. So she is able to say that for 80\% of the customers with 100\% accuracy (ie. 0\% error), when income >20k and there are 0 missed payments, the application is approved. 
}
  \label{fig:janeimages_intrees}
\end{figure}
\begin{figure}[!h]
  \centering
    \includegraphics[width=.9\textwidth]{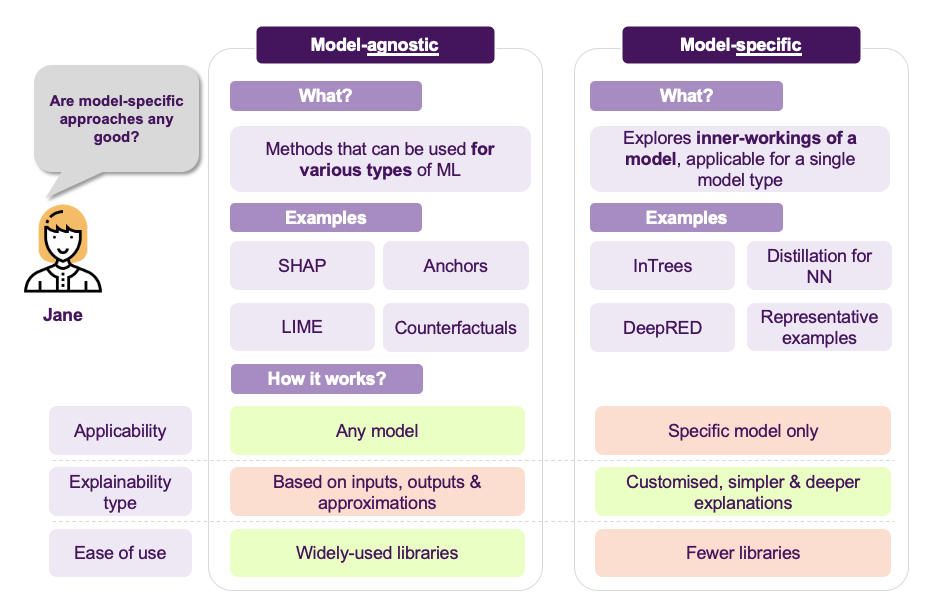}
  \caption{A short comparison of model agnostic vs model specific approaches.}
  \label{fig:janeimages_agnosticvs}
\end{figure}

Jane thinks about it and she decides to use anchors in order to achieve just that, generate easy-to-understand ``if-then'' rules that approximate the opaque model's behaviour in a local area (Figure \ref{fig:janeimages_anchors}). The resulting rules would now look something like ``if salary is greater than $20k$ \pounds ~   and there are no missed payment, then the loan is approved."

\begin{figure}[h]
  \centering
    \includegraphics[width=.9\textwidth]{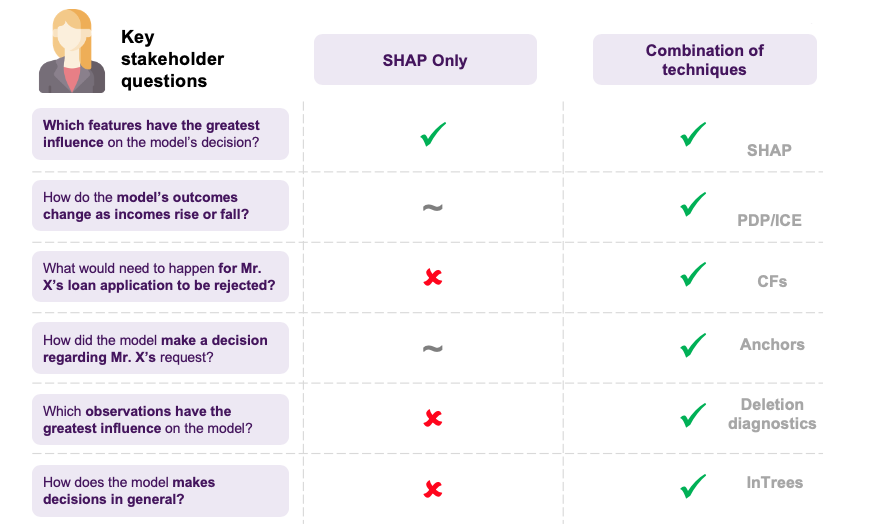}
  \caption{A list of possible questions of interest when explaining a model. This highlights the need for combining multiple techniques together and that there is no catch-all approach.}
  \label{fig:janeimages_comparison}
\end{figure}

\begin{figure}[h]
  \centering
    \includegraphics[width=.9\textwidth]{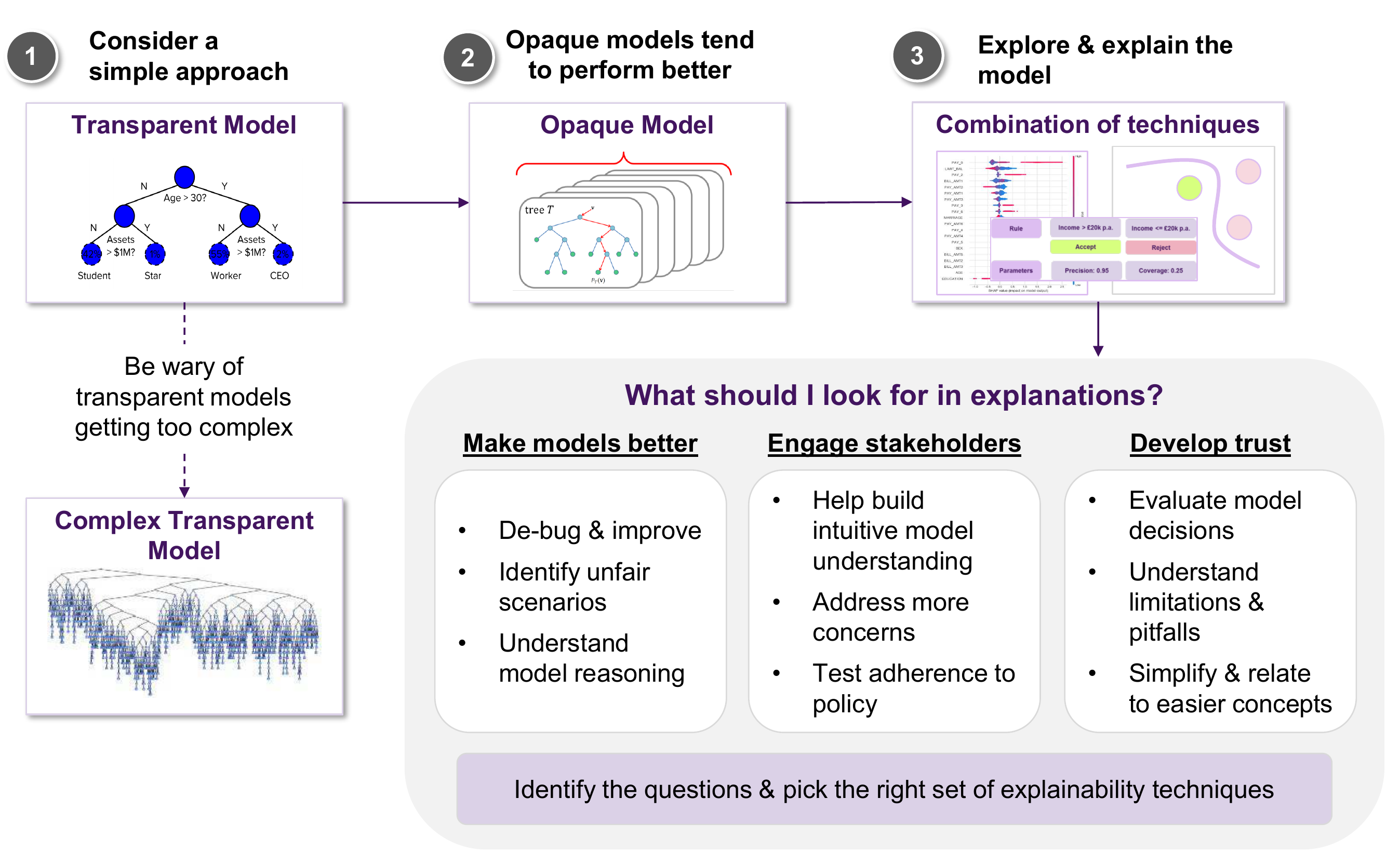}
  \caption{A sample pipeline, that is, a ``cheat sheet" of sorts for approaching explainability.}
  \label{fig:janeimages_cheat_sheet}
\end{figure}

Following these findings, the stakeholders are happy with both the model's performance and the degree of explainability. However, upon further inspection, they find out that there are some data points in the training dataset that are too noisy, probably not corresponding to actual data, but rather to instances that were included in the dateset by accident. They turn to Jane, in order to get some insights about how deleting these data points from the training dataset would affect the models behaviour. Fortunately, deletion diagnostics show that omitting these instances would not affect the models performance, while they were able to identify some points that could significantly alter the decision boundary, too (Figure \ref{fig:janeimages_deletion}). All of these helped the stakeholder understand which training data points were more influential for the model. 

\begin{figure}[!h]
  \centering
    \includegraphics[width=.9\textwidth]{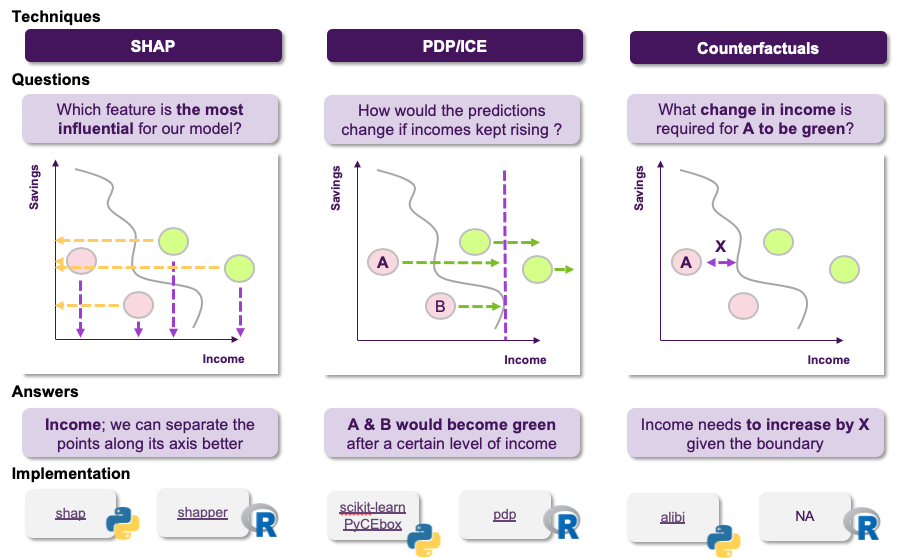}
  \caption{Using SHAP, PDF and counterfactuals, visualized in terms of instances.}
  \label{fig:janeimages_methods1}
\end{figure}
\begin{figure}[h]
  \centering
    \includegraphics[width=.9\textwidth]{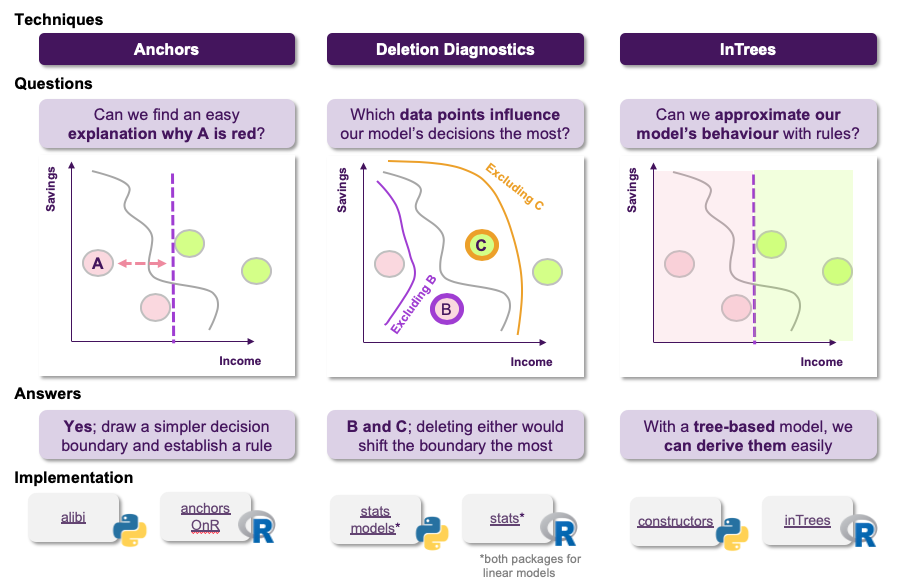}
  \caption{Using anchors, deletion diagnostics and intrees, visualized in terms of instances.}
  \label{fig:janeimages_methods2}
\end{figure}

Finally, as an extra layer of protection, the stakeholders ask Jane if it is possible to have a set of rules describing the model's behaviour on a global scale, so they can inspect it to find out whether the model has picked up any undesired functioning. At this point, Jane thinks that they should utilize the Random Forest's structure, which is an ensemble of Decision Trees. This means, that they already consist of a large number of rules, so it makes sense to go for an approach that is able to extract the more robust ones, such as inTrees (Figure \ref{fig:janeimages_intrees}).

The above example showcases how different XAI approaches can be applied to a model to answer various types of questions. Furthermore, the last point highlights an interesting distinction, as SHAP, anchors and counterfactuals that are model agnostic, while inTrees are model-specific, utilizing the model's architecture to produce explanations. There are some points to note here (cf. Figure \ref{fig:janeimages_agnosticvs}): model agnostic techniques apply  to any model, and so if benchmarking a whole range of models, inspecting their features, model agnostic methods offer consistency in interpretation. On the other hand, since these approaches have to be very flexible, a significant amount of assumptions and approximations may be made, possibly resulting in poor estimates or undesired side-effects, such as susceptibility to adversarial attacks \cite{646264}. Model-specific could also facilitate developing more efficient algorithms or custom flavoured explanations, based on the model's characteristics.

Another factor to take into consideration has to do with the libraries, since model-agnostic approaches are usually widely used and compatible with various popular libraries, whereas model-specific ones are emerging and fewer, with possibly only academic libraries being available. 
Overall, attempting to use a larger set of XAI methods allows for deeper inquiry (cf. Figure \ref{fig:janeimages_comparison}).

These insights are summarized in terms of a ``cheat sheet''. Figure \ref{fig:janeimages_cheat_sheet} discuss a sample pipeline in terms of approaching explainability for machine learning, and Figure \ref{fig:janeimages_methods1}\footnote{Links to packages (in Python and R): \tt\tiny\\ shap.readthedocs.io/en/latest/,\\  cran.r-project.org/web/packages/shapper/index.html,\\ scikit-learn.org/stable/modules/partial\_dependence.html,\\ bgreenwell.github.io/pdp/articles/pdp.html,\\    docs.seldon.io/projects/alibi/en/latest/} and Figure  \ref{fig:janeimages_methods2}\footnote{Links to packages (in Python and R): \tt\tiny \\ docs.seldon.io/projects/alibi/en/latest/,\\  github.com/viadee/anchorsOnR,\\  www.statsmodels.org/stable/generated/statsmodels.stats.outliers\_influence.OLSInfluence.html,\\  www.rdocumentation.org/packages/stats/versions/3.6.2,\\  github.com/IBCNServices/GENESIM/blob/master/constructors/inTrees.py,\\  cran.r-project.org/web/packages/inTrees/index.html} discusses possible methods.

\section{Future Directions}

\begin{figure}[h]
  \centering
    \includegraphics[width=.9\textwidth]{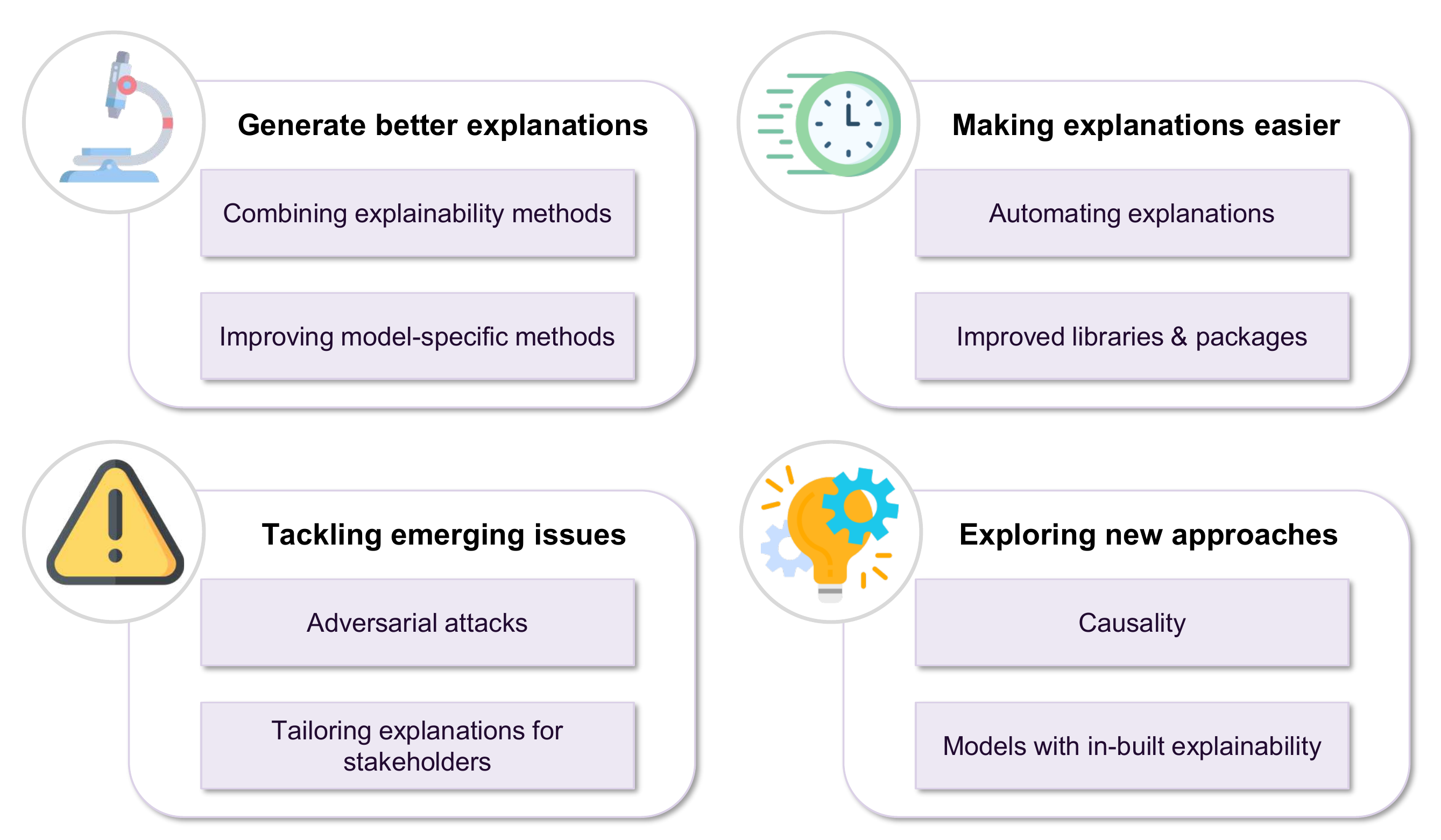}
  \caption{Possible avenues for XAI  research.}
  \label{fig:janeimages_future}
\end{figure}

This survey offers an introduction in the various developments and aspects of explainable machine learning. Having said that, XAI is a relatively new and still developing field, meaning that there are many research and operational open problems that need to be considered, as research progresses (cf.  Figure \ref{fig:janeimages_future}). 

One of the first things that comes to mind is related to the way that different explanation types fit with each other. If we take a close look at the presented approaches, we will find out that while there is some overlap between the various explanation types, for the most part they appear to be segmented, each one addressing a different question. Moreover, there seems to be no clear way of combining them in order to produce a more complete explanation. This hinders the development of pipelines that aim at automating explanations, or even reaching an agreement on how a complete explanation should look like.

On a more practical level, there are only a few XAI approaches that come with efficient implementations. This could be justified by the fact that the field is still young and emerging, but it impedes the deployment of XAI in large scale applications, nonetheless.

Another aspect that could receive more attention in the future, is developing stronger  model-specific approaches. The advantage of exploring this direction is that the resulting approaches would be able to utilize a model's distinct features in order to produce explanations, probably improving fidelity, as well as allowing to better analyze the model's inner workings, instead of just explaining its outcome. Furthermore, a side note related to the previous point is that this would probably facilitate coming up with efficient algorithmic implementations, since the new algorithms would not rely on costly approximations.

This last point leads to a broader issue that needs to be resolved, which is building trust towards the explanations themselves. As we mentioned before, recent research has showcased how a number of popular, widely used, XAI approaches are vulnerable to adversarial attacks \cite{646264}. Information like that raises questions about whether the outcome of a XAI technique should be trusted or it has been manipulated. In addition, other related issues about the fitness of some of the proposed techniques to address general explainability can be found in the literature \cite{kumar2020problems}.

Another line of research that has recently gained traction is about designing hybrid models, combining the expressiveness of opaque models with the clear semantics of transparent models, as in \cite{munkhdalai2020locally}, where linear regression is combined with neural networks, for example. This direction could not only help bridge the gap between opaque and transparent models, but could also aid the development of state-of-the-art performing explainable models.

Finally, as XAI matures, notions of causal analysis should be incorporated to new approaches \cite{miller2019explanation,pearl2018theoretical}. This is already a major driver in fundamental problems in other areas, such as fairness and bias in machine learning \cite{10.1145/2090236.2090255,NIPS2017_6995}, so we should expect it to play an integral part in the future of the XAI literature.

\clearpage

\bibliographystyle{abbrv}
% \bibliography{sample}

\end{document}